\documentclass[final]{cvpr}

\usepackage{times}
\usepackage{epsfig}
\usepackage{graphicx}
\usepackage{amsmath}
\usepackage{amssymb}

\usepackage[utf8]{inputenc} 
\usepackage[T1]{fontenc}    
\usepackage{url}            
\usepackage{booktabs}       
\usepackage{amsfonts}       
\usepackage{nicefrac}       
\usepackage{microtype}      
\usepackage{pbox}
\usepackage{epstopdf}
\usepackage{subfigure}
\usepackage{xspace}
\usepackage{comment}
\usepackage{lipsum}
\usepackage{bm}
\usepackage{bbm}
\usepackage{tabularx}
\usepackage{setspace}
\usepackage{sidecap}
\usepackage{xcolor}
\usepackage{wrapfig}
\usepackage{multirow}

\newcommand{\eric}[1]{{\color{green}[Eric: #1]}}
\newcommand{\refine}[1]{{#1}}

\newcommand{\BG}[1]{{\color{magenta}[Boqing: #1]}}

\newcommand{\bx}{\mathbf{x}}
\newcommand{\bz}{\mathbf{z}}
\newcommand{\by}{\mathbf{y}}
\newcommand{\btheta}{\boldsymbol{\theta}}

\usepackage[pagebackref=true,breaklinks=true,colorlinks,bookmarks=false]{hyperref}

\def\cvprPaperID{2326} 
\def\confYear{CVPR 2021}

\begin{document}

\title{Complete \& Label: A Domain Adaptation Approach to Semantic Segmentation of LiDAR Point Clouds}

\author{Li Yi \quad\quad Boqing Gong \quad\quad Thomas Funkhouser\\
Google Research\\
{\tt\small \{ericyi, bgong, tfunkhouser\}@google.com}
}

\maketitle

\begin{abstract}
We study an unsupervised domain adaptation problem for the semantic labeling of 3D point clouds, with a particular focus on domain discrepancies induced by different LiDAR sensors. Based on the observation that sparse 3D point clouds are sampled from 3D surfaces, we take a Complete and Label approach to recover the underlying surfaces before passing them to a segmentation network.  Specifically, we design a Sparse Voxel Completion Network (SVCN) to complete the 3D surfaces of a sparse point cloud. Unlike semantic labels, to obtain training pairs for SVCN requires no manual labeling. We also introduce local adversarial learning to model the surface prior. The recovered 3D surfaces serve as a canonical domain, from which semantic labels can transfer across different LiDAR sensors. 
Experiments and ablation studies with our new benchmark for cross-domain semantic labeling of LiDAR data show that the proposed approach provides 6.3-37.6\% better performance than previous domain adaptation methods.
\end{abstract}

\section{Introduction}
\noindent Semantic segmentation of LiDAR point clouds is important for many applications, including autonomous driving, semantic mapping, and construction site monitoring to name a few.  Given a LiDAR sweep (frame), the goal is to produce a semantic label for each point.

\begin{figure}[t]
\label{fig:teaser}
  \begin{center}
    \includegraphics[width=\linewidth]{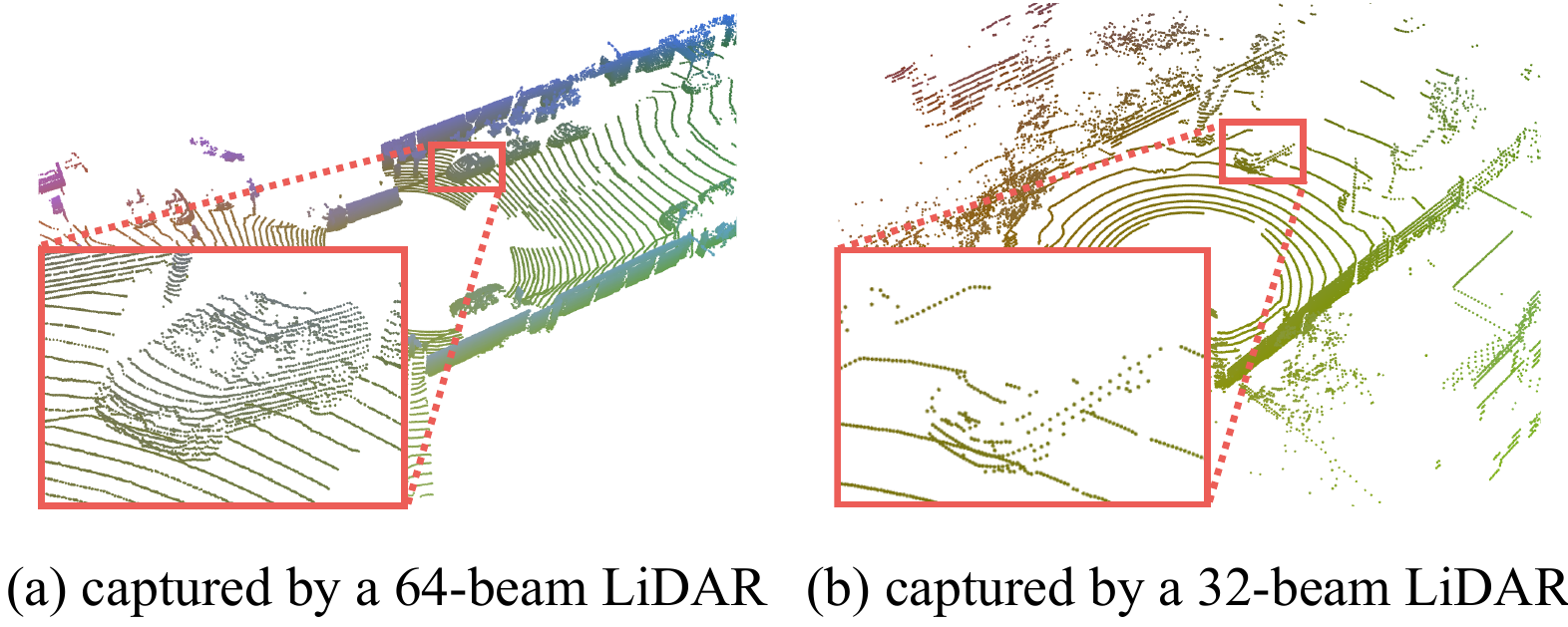}
  \end{center}
  \vspace{-0.3cm}
  \caption{The sampling discrepancy between point clouds captured by two LiDAR sensors. All figures are best viewed in color.}
  \vspace{-10pt}
\end{figure}

Although there is a great potential for deep neural networks on this semantic segmentation task, their performance is limited by the availability of labeled training data.  Acquiring manual labels for 3D points is very expensive. Several datasets have recently been released by autonomous driving companies \cite{behley2019semantickitti, caesar2019nuscenes, chang2019argoverse, geiger2013vision, geyer2020a2d2, huang2018apolloscape, lyft2019, sun2019scalability}. However, each has a different configuration of LiDAR sensors, which produce different 3D sampling patterns (Figure \ref{fig:teaser}), and each covers distinct geographic regions with distinct distributions of scene contents.
As a result, deep networks trained on one dataset do not perform well on others. 

There is a domain adaptation problem.  While the mismatch of scene contents is similar to those studied in 2D visual domain adaptation~\cite{patel2015visual,csurka2017domain}, the sampling mismatch is unique to 3D point clouds. Each time a new LiDAR sensor configuration is selected, data is acquired with a different 3D sampling pattern, so models trained on the old data are no longer effective, and new labeled data must be acquired for supervised training in the conventional machine learning paradigm. In contrast, domain adaptation aims to take better advantage of the old labeled data by revealing unlabeled data of the new LiDAR configuration to a machine learner so that it can account for the new scenarios. 

To address the sampling caused domain gap, we observe that LiDAR samples have an underlying geometric structure, and domain adaptation can be performed more effectively with a 3D model leveraging that structure.  Specifically, assuming the physical world is composed of 3D surfaces, and that LiDAR sensor samples come from those surfaces, we address the domain adaption problem by transforming it into a 3D surface completion task.  That is, if we can recover the underlying complete 3D surfaces from sparse LiDAR point samples, and train networks that operate on the completed surfaces, then we can leverage the labeled data from any LiDAR scanner to work on the data from any other.  

The motivation for this approach is that surface completion is an easier task than semantic segmentation.  First, there are strong priors on the shapes of 3D surfaces encountered in the real world, and thus a network trained to densify a point cloud can learn and leverage those priors with relatively little training data.   Second, surface completion can be learned from self-supervision (e.g., from multi-view observations) and/or from synthetic datasets (e.g., from sampled computer graphics models).  Unlike semantic segmentation, no manual labels are required.   We train our completion network with supervision from complete surfaces reconstructed from multiple frames of LiDAR data.

Our network architecture is composed of two phases: surface completion and semantic labeling.   In the first phase, we use a sparse voxel completion network (SVCN) to recover the 3D surface from a LiDAR point cloud.   In the second phase, we use a sparse convolutional U-Net to predict a semantic label for each voxel on the completed surface.   

Extensive experiments with different autonomous vehicle driving datasets verify the effectiveness of our domain adaptation approach to the semantic segmentation of 3D point clouds.  For example, using a network trained on the Waymo open dataset~\cite{sun2019scalability} to perform semantic segmentation on the nuScenes dataset~\cite{caesar2019nuscenes} provides an absolute mIoU improvement of 6.0 over state-of-the-art domain adaptation methods. Similarly, training on nuScenes and testing on Waymo provides an absolute mIoU improvement of 10.4 over prior arts.

Our contributions are three-fold. First and foremost, we identify the cross-sensor domain gap for LiDAR point clouds  caused by sampling differences, and we propose to recover complete 3D surfaces from the point clouds to eliminate the discrepancies in sampling patterns.  Second, we present a novel sparse voxel completion network, which efficiently processes sparse and incomplete LiDAR point clouds and completes the underlying 3D surfaces with high resolution. Third, we provide thorough quantitative evaluations to validate our design choices on three datasets.

\section{Related Work}
\noindent\textbf{Unsupervised domain adaptation. }  Conventional machine learning relies on the assumption that training and test sets share the same underlying distribution, but the practice often violates the assumption. Unsupervised domain adaptation (UDA)~\cite{csurka2017domain,patel2015visual} handles the mismatch by revealing some test examples  to the machine learner such that it can account for the test-time scenarios while learning from the training set. Early work on UDA mainly reweighs~\cite{shimodaira2000improving,sugiyama2008direct,zhang2013domain} or re-samples~\cite{gong2013connecting,gong2013reshaping} the source-domain examples to match the target distribution. Besides, there is a fruitful line of works on learning domain-invariant representations, such as subspace alignment~\cite{fernando2013unsupervised} and interpolation~\cite{gong2017geodesic,gopalan2011domain}, adversarial training~\cite{ganin2016domain,tzeng2017adversarial,bousmalis2017unsupervised,shrivastava2017learning,hoffman2017cycada}, maximum mean discrepancy~\cite{long2015learning}, maximum classifier discrepancy~\cite{saito2018maximum}, correlation alignment~\cite{sun2016deep}, etc. As noted in~\cite{qin2019pointdan}, these methods by design align two domains in a holistic view and fail to capture the idiosyncratic geometric properties in  point clouds. 

\noindent\textbf{Domain adaptation for 3D point clouds.} Relatively little work has been done to study domain adaptation for 3D point clouds.  Rist et al.~\cite{rist2019cross} propose that dense 3D voxels are preferable to point clouds for sensor-invariant processing of LiDAR point clouds.  Salah et al.~\cite{saleh2019domain} propose a CycleGAN approach to the adaptation of 2D bird's eye view images of LiDAR between synthetic and real domains.  Wu et al.~\cite{wu2019squeezesegv2} compensate for differences in missing points and intensities between real and synthetic data using geodesic correlation alignment.  Qin et al.~\cite{qin2019pointdan} and Wang et al.~\cite{wang2019range} propose multi-scale feature matching methods for object detection from 3D point clouds.  None of these methods explicitly account for differences in point sampling patterns in the 3D domain.

\noindent\textbf{Deep 3D semantic segmentation.} We target at deep 3D semantic segmentation in this paper, which associates semantic labels to 3D data via deep learning approaches. Different from 2D images, 3D data can be represented in various forms, introducing extra challenges for deep learning methods design. Early works use dense voxel grid to represent 3D objects and leverage dense 3D convolution to predict semantic labels~\cite{wu20153d, qi2016volumetric}, with usually a limitted resolution due to the heavy computation cost. To reduce the computation load, point cloud based methods are proposed which directly operate on point sets~\cite{qi2017pointnet, qi2017pointnet++, li2018pointcnn, su2018splatnet, thomas2019kpconv}. To further leverage the relationship among 3D points, deep neural networks working on graphs~\cite{yi2017syncspeccnn, wang2019dynamic} and meshes~\cite{bronstein2017geometric, huang2019texturenet, hanocka2019meshcnn} are used. Recently, sparse convolution based methods~\cite{graham2017submanifold, graham20183d, choy20194d} have been very popular, achieving superior performance on various indoor and outdoor semantic segmentation benchmarks. They treat 3D data as a set of sparse voxels and restrict 3D convolution to these voxels. Our segmentation backbone is based upon SparseConvNet~\cite{graham20183d} but we focus on improving its domain transfer ability to 3D data with different sampling patterns.

\begin{figure*}[t]
    \centering
    \includegraphics[width=\linewidth]{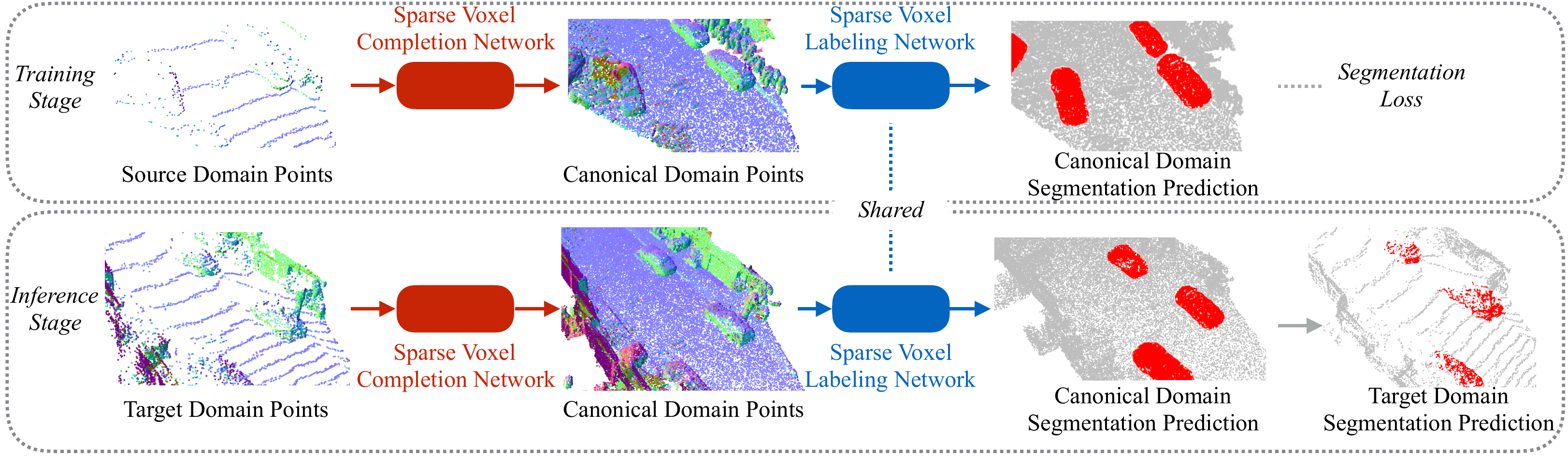}
    \vspace{-15pt}
    \caption{The overall pipeline of our ``complete and label'' approach.}
    \label{fig:pipeline}
    \vspace{-10pt}
\end{figure*}

\noindent\textbf{Deep 3D shape completion.} Deep 3D shape completion aims at completing missing geometry pieces of some partial 3D observation using deep learning methods. Dense voxel representation has been explored to complete single 3D objects~\cite{dai2017shape, yang20173d, han2017high} and indoor scenes~\cite{song2017semantic}. The heavy computation cost is a big issue for these methods, making them not scale well to large LiDAR point clouds. To improve the computation efficiency, octree-based methods have been proposed \cite{riegler2017octnetfusion, tatarchenko2017octree, zhang2018efficient} which are able to produce high resolution 3D outputs. We present a sparse voxel completion network sharing similar flavors to \cite{riegler2017octnetfusion, tatarchenko2017octree, zhang2018efficient} with an improved network architecture and loss function. We demonstrate how we could complete sparse LiDAR point clouds with high resolution using sparse convolution when the output structure is unknown and also one main difference is that we consider the application of shape completion to 3D domain adaptation.
Another relevant track of works study point cloud upsampling using deep learning methods~\cite{yu2018pu, yu2018ec, li2019pu}. They usually require an upsampling factor and have no control on the sampling patterns of the results. 

\section{Method}
This paper proposes a method to overcome the domain gap caused by different LiDAR sensors' 3D point sampling.  Observing that all the sensors acquire samples of 3D surfaces, we propose a two-stage approach, where a sensor-specific surface completion neural network first recovers the underlying 3D surfaces from the sparse LiDAR point samples, and then a sensor-agnostic semantic segmentation network assigns labels to the recovered 3D surfaces.   This two-phase approach focuses the domain adaptation problem on surface completion, which can be learned with self-supervision.

\subsection{Overview and notations}

Figure~\ref{fig:pipeline} illustrates the overall workflow of our approach.  We consider an unsupervised domain adaptation (UDA) setting, but our approach is readily applicable to other  settings such as multi-domain adaptation~\cite{multi-domain} and open domain adaptation~\cite{open-domain-1,open-domain-2}. In UDA, we have access to a set of labeled LiDAR point clouds, $\{\bx_i^s, \by_i^s\}_{i=1}^{N_s}$, from the source domain  and a set of unlabeled LiDAR point clouds $\{\bx_j^t\}_{j=1}^{N_t}$ in a target domain, where $\bx_i^s\in \mathbb{R}^{T_i^s\times 3}$ and $\bx_j^t\in \mathbb{R}^{T_j^t\times 3}$ represent two sets of $T_i^s$ and $T_j^t$ 3D points, respectively, and $\by_i^s\in\mathcal{Y}=\{1, ..., Y\}^{T_i^s}$ corresponds to a per-point semantic label ranging within $Y$ different classes. The two sets of point clouds are captured with different LiDAR sensors, which have their unique sampling patterns. Our goal is to learn a segmentation model that achieves high performance on the target-domain LiDAR points. 

To cope with the domain gap caused by different LiDAR sensors, we learn neural surface completion networks to recover the  3D surfaces underlying incomplete 3D point clouds. Denote by $\psi^s(\bx_i^s)\in\mathbb{R}^{M_i^s\times 3}$ and $\psi^t(\bx_j^t)\in\mathbb{R}^{M_j^t\times 3}$ the surface completion networks for the two domains, respectively, where $M_i^s$ and $M_j^t$ are the numbers of dense points used to represent the completed surfaces. We say the 3D surfaces reside in a \textit{canonical domain}.

We train a semantic segmentation network, $\phi(\psi^s(\bx_i^s))$, over this canonical domain by using the labeled training set of the source domain, and then apply it to the densified point clouds of the target domain, i.e., $\phi(\psi^t(\bx_i^t))$. The per-point labels of the original target-domain point cloud $\bx_i^t$ are obtained by projecting the segmentation results back to the target domain.

\subsection{SVCN for Surface Completion}
\label{sec: method_sparse_comp}
This section describes the sparse voxel completion network (SVCN), which recovers the underlying 3D surfaces from a sparse, incomplete LiDAR point cloud and is the core of our approach. 

\subsubsection{Architecture}
\begin{figure*}[t!]
    \centering
    \includegraphics[width=0.95\linewidth]{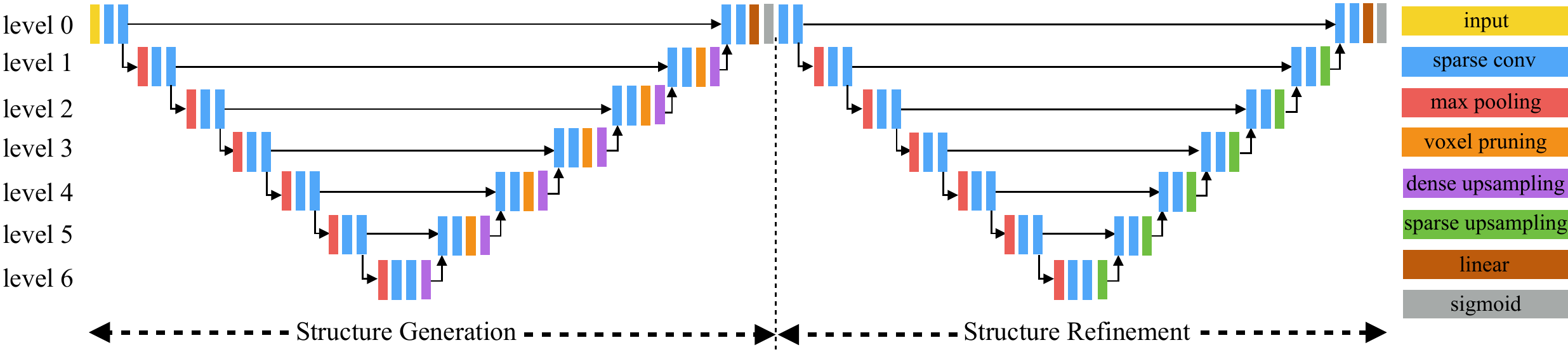}
    \vspace{-2pt}
    \caption{The architecture of the sparse voxel completion network. We use a sparse convolution with a kernel size 3 and a stride 1, max pooling with a kernel size 2 and a stride 2. Both dense and sparse upsampling are done with a factor of 2.\refine{We supervise the structure generation module and the structure refinement module with separate losses after each module as described in Section~\ref{sec:training_alg}}}
    \label{fig:architecture}
    \vspace{-10pt}
\end{figure*}

Figure~\ref{fig:architecture} shows the architecture of  SVCN, which comprises a structure generation sub-net and a structure refinement sub-net. The former consumes a  set of sparse voxels obtained by voxelizing an input point cloud, and it outputs denser voxels representing the underlying 3D surfaces from which the input points are sampled. The structure refinement network then prunes out redundant voxels.

Both sub-nets are highly relevant to the sparse convolutional U-Net~\cite{graham20183d}, which is  an encoder-decoder architecture involving a series of sparse conv/deconv operations. Multi-scale features can be integrated, and skip connections provide additional information pathways in the network. However, the sparse convolutional U-Net is not directly applicable to our setting since it applies all convolutional operations only to active sites without changing the voxel structure, while we need extrapolation.

\noindent\textbf{Structure generation network.}
In order to generate new structures for completion purposes, we replace sparse deconvolutions with dense upsampling and voxel pruning operations. Specifically, in the decoder, each voxel in the lower resolution level $l$ will generate $2^3$ voxels in the higher resolution level $l-1$ after a dense upsampling operation (see the layers in purple in Figure~\ref{fig:architecture}). Low-resolution voxel features are also duplicated to the corresponding positions of high-resolution voxels. 

The above procedure allows generating new structures, but it could easily break the inherent sparsity of voxelized 3D surfaces. Similar to \cite{riegler2017octnetfusion, tatarchenko2017octree, zhang2018efficient}, we introduce a voxel pruning module to trim voxels and avoid expanding too many in the decoder. Given a set of voxels equipped with features, the voxel pruning module applies a linear layer together with a sigmoid function on each voxel, and outputs a probability score indicating the existence of each voxel. At training time, only the ground truth voxels are kept. At test time, we prune the voxels whose existence probabilities are lower than $0.5$.

To maintain the faithfulness of the generated shape to input voxels, for each resolution level $l$, we single out intersections between the densely upsampled voxels in the decoder and the sparse voxels in the corresponding encoder level, excluding these voxels from pruning. The skip connections need  special care. Through them, we pass  encoder features to the upsampled voxels, and zeros for newly generated voxels because they have no counterparts in the encoder.

\noindent\textbf{Structure refinement network.}
The structure generation network is able to generate new structures for shape completion purposes. However, since we prune voxels using the ground truth existence probabilities on each level during training, the network could be sensitive to noisy outlier points (e.g., an outlier input voxel could possibly add a big chunk of wrong voxels to the final prediction). To cope with this issue, we introduce a structure refinement network, which is essentially a sparse convolutional U-Net adding no new voxels any more. Instead, it predicts an existence confidence score for each voxel. This is achieved by replacing the dense upsampling and voxel pruning modules in the structure generation network with sparse upsampling operations, which unpool voxel features only to the voxels that exist in the higher encoder level (see the layers in green in Figure~\ref{fig:architecture}). \refine{This way, the network is able to reevaluate the structure generation outputs across the whole scene in a spirit similar to stacked hourglass networks~\cite{newell2016stacked}.}

For more details of the SVCN network architecture, please refer to the supplementary materials.

\subsubsection{Training Data}
We need to prepare training data, $\{(\bz_i^s,\bz_i^c)\}$ and $\{(\bz_i^t,\bz_i^c)\}$, for the surface completion networks $\psi^s$ and $\psi^t$ of the source and target domains, respectively, where $\bz_i^c$ stands for a dense surface point cloud in the canonical domain from which we can sample both a source-domain point cloud $\bz_i^s$ and a target-domain point clouds $\bz_i^t$. It is important to note that the training data for the surface completion network SVCN could be different from that for semantic segmentation, so we use different notations here. Indeed, one advantage of surface completion is that it can be learned from self-supervision which does not require manual labels. Exemplar supervisions include dense  surface points via simulation, multi-view registration, and high-resolution LiDAR point clouds, to name a few. We first describe how we obtain the dense surface point clouds $\{\bz_i^c\}$, followed by how to sample domain-specific incomplete point clouds $\{\bz_i^s\}$ or $\{\bz_i^t\}$ for constructing the training pairs for SVCN.

\begin{figure}
    \centering
    \includegraphics[width=\linewidth]{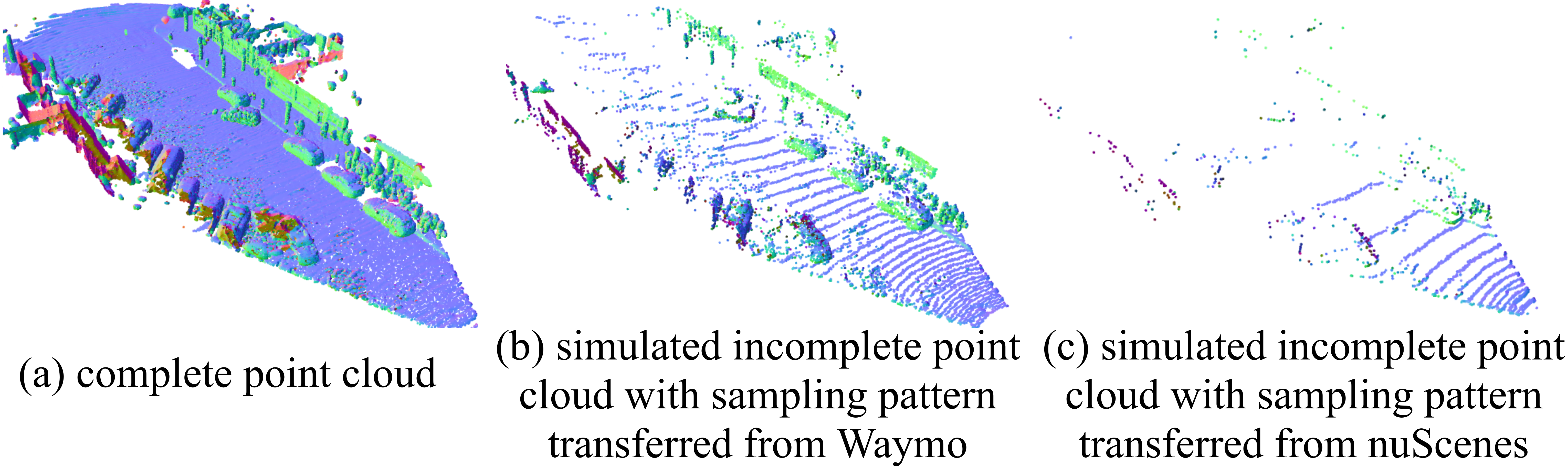}
    \vspace{-15pt}
    \caption{An example of the generated training data for SVCN.}
    \label{fig:training_data}
    \vspace{-15pt}
\end{figure}

\noindent\textbf{Dense surface point clouds.}
To obtain the dense point clouds of complete 3D surfaces, we leverage the LiDAR sequences in existing autonomous driving datasets, for example~\cite{sun2019scalability}. Specifically, we aggregate multiple LiDAR frames within a sequence to generate a denser and more complete point cloud. Poisson surface reconstruction \cite{kazhdan2006poisson} is then applied to recover the underlying mesh surfaces \refine{with a SurfaceTrimmer step removing parts with low sampling density}. We discretize a surface by uniformly sampling points on it, ensuring the point resolution is higher than the resolutions in the source or target domain. An example is shown in Figure~\ref{fig:training_data}(a). The complete scene point clouds act as a canonical domain with uniform sampling patterns. 

\noindent\textbf{Domain-specific incomplete point clouds.}
Given the dense, complete surface point clouds $\{\bz_i^c\}$, we simulate a ``virtual LiDAR'' to generate incomplete point clouds for the source (target) domain such that the virtual LiDAR point  clouds $\{\bz_i^s\}$ ($\{\bz_i^t\}$) share the same distribution as the real point clouds in that domain. In particular, we propose a polar sampling scheme to implement the ``virtual LiDAR''. First, we randomly pick up a reference point cloud from a domain and compute the polar coordinate $(r, \theta, \phi)$ for each point $(x, y, z)$, where $r=\sqrt{x^2+y^2+z^2}$, $\theta=\text{atan2}(\sqrt{x^2+y^2},z)$, $\phi=\text{atan2}(y, x)$. We argue that $(\theta, \phi)$ reveals the sampling pattern in this point cloud without much scene-specific information and can be used to re-sample a different complete scene point cloud to simulate the corresponding sampling pattern. Second, we select a sensor location in the complete scene point cloud, remove occluded points, and convert the rest points into their polar coordinates. Finally, we  search for the nearest neighbor point in the complete scene point cloud for each point in a reference frame and sample these points to imitate the reference sampling pattern. 
Notice this is done in the $(\theta, \phi)$ space to transfer the sampling pattern only. In Figure~\ref{fig:training_data} (b) and (c), we shown the simulated incomplete point clouds with sampling patterns transferred from reference point clouds in the Waymo \cite{sun2019scalability} and nuScenes \cite{caesar2019nuscenes} datasets, respectively.

\subsubsection{Training Algorithm}
\label{sec:training_alg}

Given the paired training data, we convert them to voxels and employ a voxel-wise binary cross-entropy loss to first pre-train the structure generation sub-net. We then fix it and switch it to the inference mode, using the predicted voxel existence probability as the input to train the structure refinement sub-net with another voxel-wise binary cross-entropy loss. 

\noindent\textbf{Local Adversarial Learning.} Since we have a strong prior that voxels densified by SVCN should lie on 3D surfaces, we propose an adversarial loss to capture this prior, in a similar spirit to~\cite{ledig2017photo, wang2018esrgan, li2019pu}. This loss can be added to the training of either the structure generation sub-net or the refinement sub-net. A notable property of our adversarial loss is that it is imposed over local surface patches, as opposed to the global output of SVCN. Please refer to the supplementary materials for more details.

\subsection{Segmentation in the Canonical Domain}
\label{sec: method_seg}
We train a semantic segmentation network $\phi(\cdot)$ over the canonical domain using the labeled data $\{(\bx_i^s,\by_i^s)\}$ in the source domain. We leverage a state-of-the-art 3D semantic segmentation method, MinkowskiNet~\cite{choy20194d}, as our segmentation network. Given a test point cloud $\bx_i^t$ in the target domain, we first map it to the canonical domain by the surface completion network SVCN $\psi^t(\bx_i^t)$, apply the segmentation network over it $\phi(\psi^t(\bx_i^t))$, and finally project the segmentation results back to the original target-domain point cloud $\bx_i^t$. Please refer to the supplementary material for how to propagate the source-domain labels to the dense, complete point clouds in the canonical domain and how to project segmentation results to the target domain. Both depend on nearest neighbor search and majority voting.

\section{Experiments}

We experiment with three autonomous driving datasets captured by different LiDAR configurations.
\begin{itemize}   \setlength\itemsep{-3pt}
\vspace*{-2mm}
    \item Waymo open dataset~\cite{sun2019scalability}: It contains LiDAR point cloud sequences from 1K scenes, each sequence containing about 200 frames. There are five LiDAR sensors. We use the top 64-beam LiDAR in our experiments. The LiDAR frames are labeled with 3D object bounding boxes in four categories, from which we crop the LiDAR point clouds to obtain per-point semantic labels. The data is officially split into 798 training scenes and 202 validation scenes. Following this slit, we  have  \textasciitilde160K training frames and \textasciitilde40K validation frames.
    \item nuScenes-lidarseg dataset~\cite{caesar2019nuscenes}: It contains \textasciitilde40K LiDAR frames annotated with per-point semantic labels from 1K scenes. Officially these points are cast into 16 categories for the semantic segmentation task with one additional ``ignored'' class excluded from evaluations. Different from the Waymo open dataset, it adopts a 32-beam LiDAR sensor with different configurations, causing a sampling gap from the Waymo point clouds. Following the dataset's recommendation, we train our models using \textasciitilde28K frames from 700 training scenes and evaluate on \textasciitilde6K frames from 150 validation scenes. 
    \item SemanticKITTI dataset~\cite{behley2019semantickitti, geiger2013vision}: It adopts a Velodyne 64-beam LiDAR similar to Waymo but with a different sensor configuration. Points are classified into 19 categories with one additional ``ignored'' class excluded from evaluations. Following the official recommendation, we use sequence 00-07 and 09-10 for training and evaluate on sequence 08, resulting in \textasciitilde19K training frames and \textasciitilde4K frames for evaluation.
    \vspace{-5pt}
\end{itemize}
While transferring semantic segmentation from Waymo to nuScenes-lidarseg and SemanticKITTI or the inverse, we consider the only two overlapping categories in all these three datasets: vehicles and pedestrians. The two classes are both common and safety-critical in self-driving scenes. For the domain transfer between nuScenes-lidarseg and SemanticKITTI, we consider all the 10 overlapping categories between the two datasets: car, bicycle, motorcycle, truck, other vehicle, pedestrian, drivable surface, sidewalk, terrain, and vegetation. We carefully remap the semantic categories to guarantee the class definitions in different datasets are consistent. Please refer to the supplementary material for the remapping process. The three datasets provide an organic, large-scale testbed to study domain adaptation methods for 3D point clouds. By design, our approach copes with the domain discrepancy among the three datasets caused by different configurations of LiDAR sensors.

\subsection{Sparse LiDAR Point Cloud Completion}
\label{sec:exp_comp}
We first evaluate our sparse voxel completion network (SVCN)  in this section. 

\noindent\textbf{Training data.} SVCN takes as input an incomplete point cloud and predicts its underlying complete 3D surfaces in a dense volumetric form. To generate data pairs for training and evaluation, we aggregate multiple frames within each sequence from the Waymo open dataset,  resulting in 2400 complete scene point clouds for training and 200 for test. We then sample incomplete point clouds via the ``virtual LiDAR'' described in Section~\ref{sec: method_sparse_comp}. We voxelize the complete point clouds using a voxel size of 20cm to provide ground truth supervision for various learning methods.


\noindent\textbf{Evaluation.} We use two evaluation metrics. One is voxel-level intersection over union (IoU). The other is the Chamfer Distance (CD) between the predicted voxel set and the ground truth voxel set. To compute the CD between two voxel sets, we convert each voxel set into a point cloud by keeping the center of each voxel and then compute CD between the two point clouds. 

\noindent\textbf{SVCN vs.\ Baselines.}
There is not much prior work on using sparse convolution to complete sparse LiDAR inputs. \refine{The closest baseline is ESSCNet \cite{zhang2018efficient} which achieves state-of-the-art results for semantic indoor scene completion. Our structure generation network (the first half of SVCN) is a variant of ESSCNet with two improvments. It leverages only one group in the spatial group convolution for higher generation quality, and we densely supervise voxel pruning at each resolution with resolution balancing weights. So we will also refer to the ablation baseline of our SVCN without the structure refinement as $\text{ESSCNet}^{++}$.}
A second ablation baseline is our full SVCN network trained without the local adversarial loss.


\begin{table}[h]\small
\centering
\caption{Comparison for sparse LiDAR point cloud completion (W: Waymo, N: nuScenes-lidarseg, K: SemanticKITTI).}
\label{tab:sparse_voxel_completion}
\newcolumntype{Y}{>{\centering\arraybackslash}X}
{
\setlength{\tabcolsep}{0.2em}
\begin{tabularx}{\columnwidth}{Y|Y|Y|Y|Y}
\toprule
   Test domain & Metric & $\text{ESSCNet}^{++}$ \cite{zhang2018efficient} & SVCN w/o adv. & SVCN\\
    \hline
   \multirow{2}{*}{W}&$\text{IoU}(\%)\uparrow$ & 44.1 & 46.3 & \textbf{47.5} \\
   &$\text{CD}(\text{m})\downarrow$ & 1.070 & 1.013 & \textbf{0.968}\\
    \hline
  \multirow{2}{*}{N} &$\text{IoU}(\%)\uparrow$ & 24.9 & 26.7 & \textbf{28.8} \\
   &$\text{CD}(\text{m})\downarrow$ & 1.745 & 1.730 & \textbf{1.610}\\
   \hline
  \multirow{2}{*}{K} &$\text{IoU}(\%)\uparrow$ & 40.9 & 42.9 & \textbf{44.3}  \\
   &$\text{CD}(\text{m})\downarrow$ & 1.147 & 1.122 & \textbf{1.052} \\
\bottomrule
\end{tabularx}
}
\vspace{-5pt}
\end{table}

\begin{figure}
\vspace{0pt}
    \centering
    \includegraphics[width=\linewidth]{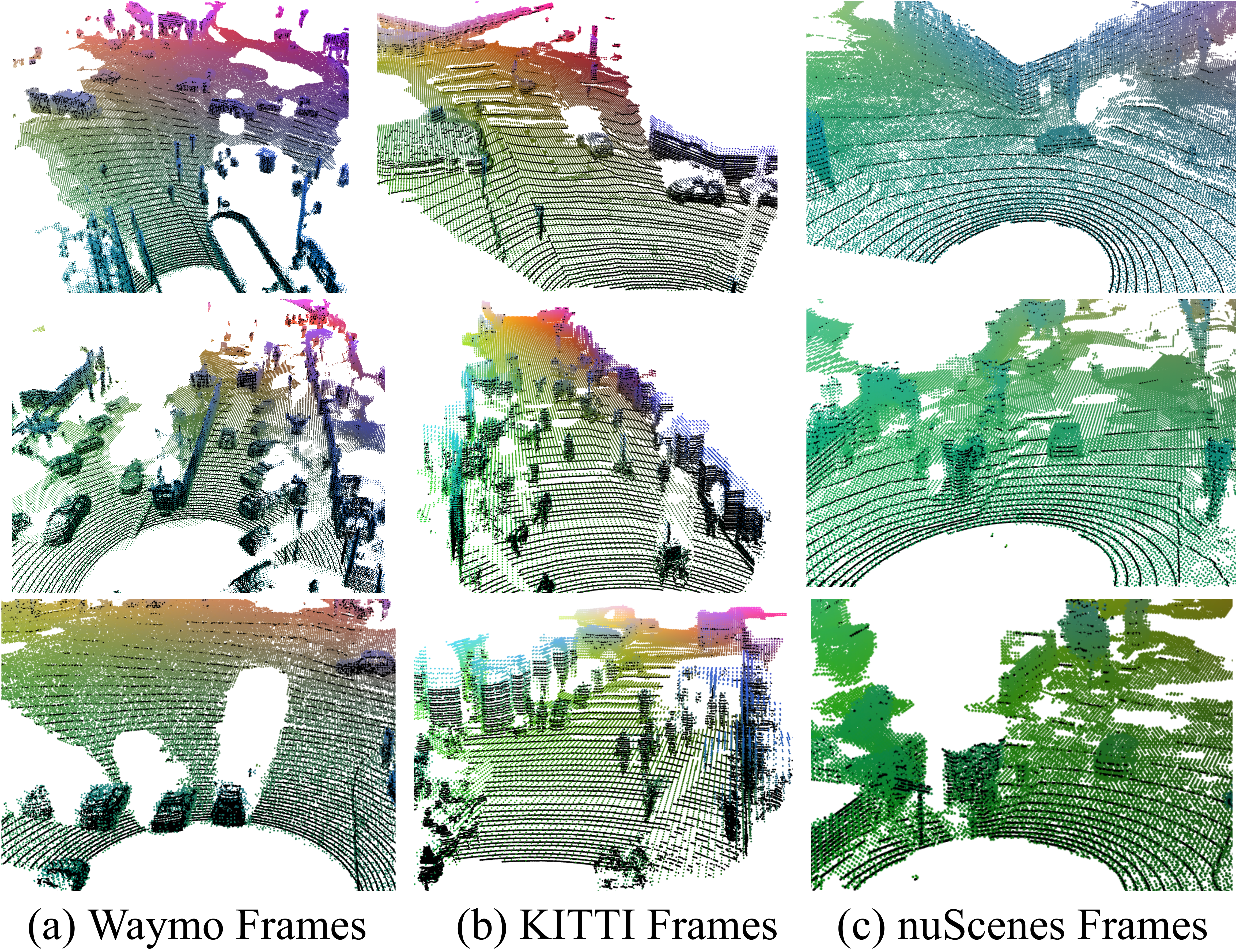}
    \vspace{-15pt}
    \caption{Visualizations of the surface completion results from different datasets. Black points indicate the original sparse incomplete LiDAR points, and we use colored points to represent the output of our sparse voxel completion network.}
    \label{fig:surf_comp_vis}
    \vspace{-10pt}
\end{figure}

\noindent\textbf{Results.} Table~\ref{tab:sparse_voxel_completion} shows the comparison results.  Our full network with local adversarial learning outperforms all competing methods for the inputs with the sampling patterns of all three datasets. Comparing $\text{ESSCNet}^{++}$ and SVCN without the local adversarial loss, we can see that the structure refinement network does improve the scene completion quality. Finally, the local adversarial loss, which accounts for surface priors,  results in better completions than SVCN without it. Notice that the LiDAR point clouds from nuScenes-lidarseg hold a much sparser sampling pattern compared with those from Waymo or SemanticKITTI, and are thus more challenging for the completion task. This is revealed by the relatively low IoU and high CD scores when the inputs hold nuScenes-lidarseg sampling patterns.

To better understand how our surface completion network SVCN could canonicalize different sampling patterns and therefore mitigate the corresponding domain gap, we visualize the surface completion results from different datasets in Figure~\ref{fig:surf_comp_vis}. We use black points to represent the incomplete LiDAR inputs and colored points for the outputs of our SVCN. It is clear that SVCN is able to recover the underlying surfaces regardless of the input sampling patterns. SVCN is also able to fill small holes to make the geometry more complete. Comparing the vehicles from Waymo and nuScenes-lidarseg datasets respectively, we observe a clear domain gap in the inputs. In contrast, after surface completion they share more similar sampling patterns and geometry.

\begin{figure*}
\vspace{0pt}
    \centering
    \includegraphics[width=0.95\linewidth]{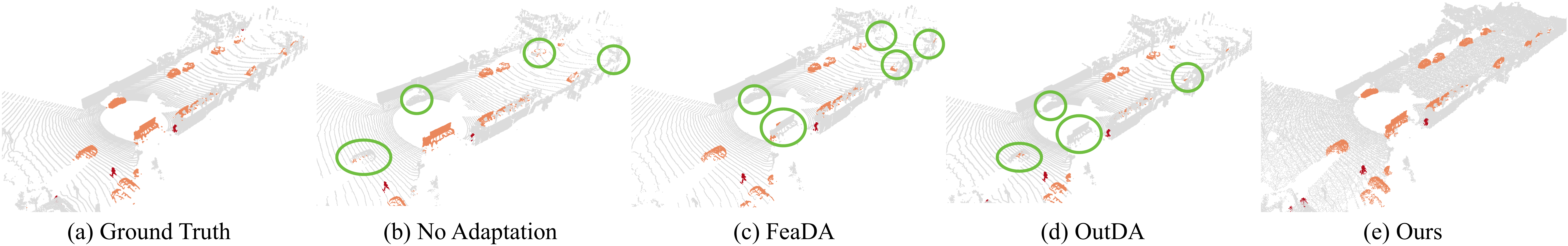}
    \vspace{-2pt}
    \caption{A comparison of different domain adaptation methods on an example Waymo frame. We consider the domain adaptation direction from nuScenes-lidarseg to Waymo dataset. Different colors indicate different semantic classes. FeaDA and OutDA represent feature space domain and output space domain adaptation respectively. We use green circles to highlight the prediction errors.}
    \label{fig:seg_results}
    \vspace{-5pt}
\end{figure*}

\subsection{Unsupervised Domain Adaptation Results}
In this section, we study the domain transfer ability of our approach among the Waymo, nuScenes-lidarseg and SemanticKITTI datasets.
We compare our method with the previous state-of-the-art method of SqueezeSegV2 (SQSGV2) on this topic~\cite{wu2019squeezesegv2}, which projects LiDAR point clouds to form 2D range images and uses 2D convolutional neural networks as the segmentation backbone. Since domain adaptation methods for 3D LiDAR point cloud segmentation has not been studied much previously, we also compare with state-of-the-art adaptation methods for 2D semantic segmentation, including feature space adversarial domain adaptation (FeaDA)~\cite{chen2017no}, output space adversarial domain adaptation (OutDA)~\cite{tsai2018learning}, and Sliced Wasserstein Discrepancy-based domain adaptation (SWD)~\cite{lee2019sliced}. In addition, we incorporate the geodesic correlation alignment technique used in~
\cite{wu2019squeezesegv2} with our 3D segmentation backbone, forming another baseline named 3DGCA. We report the results in Table~\ref{tab:sem_seg}. 

\begin{table*}[h]\small
\centering
\caption{Unsupervised domain adaptation for 3D semantic segmentation among the Waymo, nuScenes-lidarseg and SemanticKITTI datasets. N denotes nuScenes-lidarseg, W denotes Waymo and K denotes SemanticKITTI. We report the mIoU in each cell.}
\label{tab:sem_seg}
\begin{tabular}{c|c|c|c|c|c|c|c|c}
\toprule
Source$\rightarrow$Target & \#classes & No DA & FeaDA~\cite{chen2017no} & OutDA~\cite{tsai2018learning} & SWD~\cite{lee2019sliced} & SQSGV2~\cite{wu2019squeezesegv2} & 3DGCA & Ours \\ \midrule
N$\rightarrow$N & 2 & 69.9 & - & - & - & - & - & - \\ 
W$\rightarrow$N & 2 & 42.9      & 44.2      & 40.2      & 42.4    & 14.2    & 43.85      & \textbf{50.2}     \\ \midrule
N$\rightarrow$N & 10 & 54.4 & - & - & - & - & - & - \\ 
K$\rightarrow$N & 10 & 27.9 & 27.2 &	26.5 & 27.7 & 10.1 & 27.4 & \textbf{31.6}\\ \midrule
W$\rightarrow$W & 2 & 86.3 & - & - & - & - & - & - \\ 
N$\rightarrow$W & 2 & 46.2      & 48.7      & 43.6      & 49.3    & 30.4       & 48.7      & \textbf{59.7}     \\  
K$\rightarrow$W & 2 & 46.3   & 43.9       & 43.4      & 47.2    & 34.2    & 46.1      & \textbf{52.0}     \\ \midrule
K$\rightarrow$K & 10 & 50.2 & - & - & - & - & - & - \\ 
N$\rightarrow$K & 10 & 23.5 & 21.4 & 22.7 & 24.5 & 13.4 & 23.9 & \textbf{33.7}\\ \midrule
K$\rightarrow$K & 2 & 61.0 & - & - & - & - & - & - \\ 
W$\rightarrow$K & 2 & 55.0      & 56.4      & 54.1      & 56.8    & 36.8    & 56.2      & \textbf{60.4}     \\ \bottomrule
\end{tabular}
\vspace{-10pt}
\end{table*}

From Table~\ref{tab:sem_seg}, we can see an obvious performance drop when transferring segmentation networks from one domain to another. For example, compared with both training and testing on nuScenes-lidarseg (N$\rightarrow$N), training on the Waymo dataset while evaluating on the nuScenes-lidarseg dataset (W$\rightarrow$N) would cause the mean IoU (mIoU) to drop from 69.9$\%$ to 42.9$\%$. This shows the importance of studying the domain adaptation problem. Our method successfully brings the mIoU to 50.2$\%$ and outperforms the prior arts. We draw the same observation on the other pairs of domains tested.

The 2D domain adaptation methods  do not work well on the 3D point clouds. FeaDA tries to bring close two domains in a global feature space, but it fails to model rich local cues in 3D point clouds, such as sampling patterns, surfaces, and contexts. OutDA fails in most cases with no surprise because it assumes that the segmentation masks of two domains are indistinguishable. While this assumption works for 2D scenes, it breaks given different 3D sampling patterns in two domains. SWD in general does a better job than the previous two methods. However, since it does not explicitly model the cause of the domain gap, we do not observe a huge performance improvement either. The performance of SQSGV2 is significantly lower than the other methods when evaluated on 3D point clouds. It requires projecting LiDAR point clouds onto range images to exploit 2D convolutional methods. When projecting the predicted labels back to 3D, we observe huge errors especially along object boundaries, leading to the subpar performance.
The performance improvement by 3DGCA is also restricted.



\noindent\textbf{Qualitative Results.}
Figure~\ref{fig:seg_results} shows some qualitative results of both surface completion and semantic segmentation when adapting from nuScenes-lidarseg to Waymo. We can see that the baseline methods mislabel objects that are both close to and distant from the sensor location. The sparsity of distant objects is a great challenge for all methods. Our approach completes the underlying 3D surfaces from only sparse observations, making it easier for the segmentation network to handle those distant objects.

\noindent\textbf{Correlation between scene completion and domain adaptation.}
We provide additional ablation studies about the correlation between the quality of scene completion  and the performance of domain adaptation. 
We replace SVCN in our method with its variants described in Section~\ref{sec:exp_comp} and  report the resulting segmentation results in Table~\ref{tab:ablation_cal}. 
It can be clearly seen that better scene completion qualities lead to better domain transfer performances.

\begin{table}[h]\small
\centering
\caption{The segmentation mIoU of our approach when using different scene completion methods.}
\label{tab:ablation_cal}
\newcolumntype{Y}{>{\centering\arraybackslash}X}
{
\setlength{\tabcolsep}{0.2em}
\begin{tabularx}{\columnwidth}{Y|Y|Y|Y}
\toprule
    Source$\rightarrow$ Target & Ours w/o refinement & Ours w/o adv. & Ours-full \\
    \hline
    W$\rightarrow$N & 46.6 & 49.1 & \textbf{50.2}\\
    N$\rightarrow$W & 58.2  & 59.0 & \textbf{59.8}\\
\bottomrule
\end{tabularx}
}
\vspace{-10pt}
\end{table}

\noindent\textbf{Comparison with handcrafted baselines.} The sampling patterns differ across the autonomous driving datasets due to  complex interactions of various factors, such as the heights, poses, and types of LiDAR sensors, making it hard to design analytical solutions to align them. To show the difficulty of the problem and demonstrate the necessity of our approach, we also study some analytical methods aiming to align different sampling patterns. Consider the Waymo open dataset captured by 64-beam LiDARs and nuScenes consisting of 32-beam LiDAR data. We design two handcrafted baselines to align them: B1) analytically downsampling or upsampling the LiDAR beams and B2) piecewise linear interpolation. For both baselines, we first convert a LiDAR point cloud into a $H\times W$ range image, where $H$ equals the beam number of the LiDAR and $W$ is set to 2048. 
We say two points are adjacent if they have the same column index and their row indices are off by one. In B1, we simply remove the points of every other row in the range image to downsample Waymo point clouds. To upsample nuScenes, we add the midpoint of every pair of adjacent points. In  B2, we linearly interpolate every pair of adjacent points, densely adding points to the line segment and making their spacing follow a hyperparameter $\delta$. We use the same $\delta$ while densifying Waymo and nuScenes to bring their sampling patterns close. 

We compare our approach with the above two baselines in Table~\ref{tab:simp_baselines}. Our method outperforms both by large margins. B1 fails because the sampling pattern difference is on more than just the number of LiDAR beams. B2 densifies the point cloud by interpolation, but it produces ``phantom'' points along back-projections of silhouette boundaries and does not complete occluded regions. This study shows the importance of our learning-based approach and the SVCN network.

\begin{table}[h]\small
\centering
\caption{Comparison with handcrafted sampling aligning baselines.}
\label{tab:simp_baselines}
\newcolumntype{Y}{>{\centering\arraybackslash}X}
{
\setlength{\tabcolsep}{0.2em}
\begin{tabularx}{\columnwidth}{Y|Y|Y|Y|Y}
\toprule
     Src$\rightarrow$Tgt & No DA& B1 & B2 & Ours \\
    \hline
    W$\rightarrow$N & 42.9 & 46.0 & 45.1 & \textbf{50.2}\\
    \hline
    N$\rightarrow$W & 46.2 & 55.6 & 55.2 & \textbf{59.8}\\
\bottomrule
\end{tabularx}
}
\vspace{-10pt}
\end{table}

\subsection{Domain Generalization Results}
\label{sec:notarget}
Getting rid of the dependency on target domain data during training was argued to be an important feature in real applications~\cite{yue2019domain}, which allows domain generalization to multiple unseen target domains. In this section, we demonstrate our approach performs well for domain generalization.

We choose the Waymo dataset as our source domain and aim at generalizing a segmentation neural network to  nuScenes-lidarseg and SemanticKITTI  without accessing them during training. 
Our strategy is to train a generic surface completion network. For this purpose, we introduce data augmentation while generating virtual LiDAR point clouds $\{\bz_i^s\}$ from complete surface point clouds $\{\bz_i^c\}$. 
More concretely, we use $(\theta,\phi)$ to denote the polar coordinates of a Waymo point cloud and we evenly quantize $\theta$ into 64 bins. We randomly select $30\%$ to $70\%$ of the bins and use the corresponding points as our augmented reference sampling pattern. This augmentation strategy enforces SVCN to handle various sampling patterns, therefore generalizing to different target domains.
After training our SVCN using only the reference point clouds from Waymo (with the data augmentation strategy), we evaluate its surface completion quality for the nuScenes-lidarseg and SemanticKITTI sampling patterns. The evaluation metric is the same as in Section~4.1. Furthermore, we show how the surface completion quality contributes to the domain generalization performance of the semantic segmentation task using mIoU as the evaluation metric.  Table~\ref{tab:notarget} shows the results, where we additionally report the domain adaptation results for comparison (i.e., by training a target-domain-specific SVCN). 

\begin{table}[h]
\vspace{-2pt}
\centering
\small
\caption{Domain generalization from Waymo to nuScenes-lidarseg and SemanticKITTI by training a generic SVCN.
}
\label{tab:notarget}
\newcolumntype{Y}{>{\centering\arraybackslash}X}
{
\setlength{\tabcolsep}{0.2em}
\begin{tabularx}{\columnwidth}{Y|Y|Y|Y}
\toprule
    Source$\rightarrow$Target & Method & Surface Completion IoU$(\%)$/CD(\text{m}) & Semantic Segmentation mIoU($\%$)\\
    \hline
    & No Adaptation & -/- & 42.9\\
    W$\rightarrow$N  & Adaptation & 28.8/1.610 & 50.2\\
    & Generalization & 25.7/1.800 & 49.8 \\
    \hline
    & No Adaptation & -/- & 55.0\\
    W$\rightarrow$K  & Adaptation & 44.3/1.052 & 60.4 \\
    & Generalization & 42.8/1.115 & 59.6 \\
\bottomrule
\end{tabularx}
}
\vspace{-10pt}
\end{table}

We can see that generic SVCN trained under the domain generalization setting  performs a little worse on surface completion and slightly degrades the semantic segmentation results on nuScenes-lidarseg and SemanticKITTI, compared with the target-domain-specific SVCNs learned under the domain adaptation setting. However, both adaptation and generalization are better than the no adaptation baselines by  large margins, indicating the efficacy of our method.

\section{Conclusion}
In this paper, we present ``complete and label'', a novel domain adaptation approach designed to overcome the domain gap in 3D point clouds acquired with different LiDAR sensors. We show that by leveraging geometric priors, we can transform this domain adaptation problem into a 3D surface completion task, and then perform downstream tasks such as semantic segmentation on the completed 3D surfaces with sensor-agnostic networks. Extensive experiments with multiple autonomous driving datasets demonstrate the significant improvement of our approach over prior arts.


{\small
\bibliographystyle{ieee_fullname}
\bibliography{egbib}

\begin{thebibliography}{10}\itemsep=-1pt

\bibitem{behley2019semantickitti}
Jens Behley, Martin Garbade, Andres Milioto, Jan Quenzel, Sven Behnke, Cyrill
  Stachniss, and Jurgen Gall.
\newblock Semantickitti: A dataset for semantic scene understanding of lidar
  sequences.
\newblock In {\em Proceedings of the IEEE International Conference on Computer
  Vision}, pages 9297--9307, 2019.

\bibitem{bousmalis2017unsupervised}
Konstantinos Bousmalis, Nathan Silberman, David Dohan, Dumitru Erhan, and Dilip
  Krishnan.
\newblock Unsupervised pixel-level domain adaptation with generative
  adversarial networks.
\newblock In {\em Proceedings of the IEEE conference on computer vision and
  pattern recognition}, pages 3722--3731, 2017.

\bibitem{bronstein2017geometric}
Michael~M Bronstein, Joan Bruna, Yann LeCun, Arthur Szlam, and Pierre
  Vandergheynst.
\newblock Geometric deep learning: going beyond euclidean data.
\newblock {\em IEEE Signal Processing Magazine}, 34(4):18--42, 2017.

\bibitem{caesar2019nuscenes}
Holger Caesar, Varun Bankiti, Alex~H Lang, Sourabh Vora, Venice~Erin Liong,
  Qiang Xu, Anush Krishnan, Yu Pan, Giancarlo Baldan, and Oscar Beijbom.
\newblock nuscenes: A multimodal dataset for autonomous driving.
\newblock {\em arXiv preprint arXiv:1903.11027}, 2019.

\bibitem{chang2019argoverse}
Ming-Fang Chang, John Lambert, Patsorn Sangkloy, Jagjeet Singh, Slawomir Bak,
  Andrew Hartnett, De Wang, Peter Carr, Simon Lucey, Deva Ramanan, et~al.
\newblock Argoverse: 3d tracking and forecasting with rich maps.
\newblock In {\em Proceedings of the IEEE Conference on Computer Vision and
  Pattern Recognition}, pages 8748--8757, 2019.

\bibitem{chen2017no}
Yi-Hsin Chen, Wei-Yu Chen, Yu-Ting Chen, Bo-Cheng Tsai, Yu-Chiang Frank~Wang,
  and Min Sun.
\newblock No more discrimination: Cross city adaptation of road scene
  segmenters.
\newblock In {\em Proceedings of the IEEE International Conference on Computer
  Vision}, pages 1992--2001, 2017.

\bibitem{choy20194d}
Christopher Choy, JunYoung Gwak, and Silvio Savarese.
\newblock 4d spatio-temporal convnets: Minkowski convolutional neural networks.
\newblock In {\em Proceedings of the IEEE Conference on Computer Vision and
  Pattern Recognition}, pages 3075--3084, 2019.

\bibitem{csurka2017domain}
Gabriela Csurka.
\newblock {\em Domain adaptation in computer vision applications}, volume~2.
\newblock Springer, 2017.

\bibitem{dai2017shape}
Angela Dai, Charles Ruizhongtai~Qi, and Matthias Nie{\ss}ner.
\newblock Shape completion using 3d-encoder-predictor cnns and shape synthesis.
\newblock In {\em Proceedings of the IEEE Conference on Computer Vision and
  Pattern Recognition}, pages 5868--5877, 2017.

\bibitem{multi-domain}
Lixin Duan, Ivor~W Tsang, Dong Xu, and Tat-Seng Chua.
\newblock Domain adaptation from multiple sources via auxiliary classifiers.
\newblock In {\em Proceedings of the 26th Annual International Conference on
  Machine Learning}, pages 289--296, 2009.

\bibitem{fernando2013unsupervised}
Basura Fernando, Amaury Habrard, Marc Sebban, and Tinne Tuytelaars.
\newblock Unsupervised visual domain adaptation using subspace alignment.
\newblock In {\em Proceedings of the IEEE international conference on computer
  vision}, pages 2960--2967, 2013.

\bibitem{ganin2016domain}
Yaroslav Ganin, Evgeniya Ustinova, Hana Ajakan, Pascal Germain, Hugo
  Larochelle, Fran{\c{c}}ois Laviolette, Mario Marchand, and Victor Lempitsky.
\newblock Domain-adversarial training of neural networks.
\newblock {\em The Journal of Machine Learning Research}, 17(1):2096--2030,
  2016.

\bibitem{geiger2013vision}
Andreas Geiger, Philip Lenz, Christoph Stiller, and Raquel Urtasun.
\newblock Vision meets robotics: The kitti dataset.
\newblock {\em The International Journal of Robotics Research},
  32(11):1231--1237, 2013.

\bibitem{geyer2020a2d2}
Jakob Geyer, Yohannes Kassahun, Mentar Mahmudi, Xavier Ricou, Rupesh Durgesh,
  Andrew~S Chung, Lorenz Hauswald, Viet~Hoang Pham, Maximilian M{\"u}hlegg,
  Sebastian Dorn, et~al.
\newblock A2d2: Audi autonomous driving dataset.
\newblock {\em arXiv preprint arXiv:2004.06320}, 2020.

\bibitem{gong2013connecting}
Boqing Gong, Kristen Grauman, and Fei Sha.
\newblock Connecting the dots with landmarks: Discriminatively learning
  domain-invariant features for unsupervised domain adaptation.
\newblock In {\em International Conference on Machine Learning}, pages
  222--230, 2013.

\bibitem{gong2013reshaping}
Boqing Gong, Kristen Grauman, and Fei Sha.
\newblock Reshaping visual datasets for domain adaptation.
\newblock In {\em Advances in Neural Information Processing Systems}, pages
  1286--1294, 2013.

\bibitem{gong2017geodesic}
Boqing Gong, Kristen Grauman, and Fei Sha.
\newblock Geodesic flow kernel and landmarks: Kernel methods for unsupervised
  domain adaptation.
\newblock In {\em Domain Adaptation in Computer Vision Applications}, pages
  59--79. Springer, 2017.

\bibitem{goodfellow2014generative}
Ian Goodfellow, Jean Pouget-Abadie, Mehdi Mirza, Bing Xu, David Warde-Farley,
  Sherjil Ozair, Aaron Courville, and Yoshua Bengio.
\newblock Generative adversarial nets.
\newblock In {\em Advances in neural information processing systems}, pages
  2672--2680, 2014.

\bibitem{gopalan2011domain}
Raghuraman Gopalan, Ruonan Li, and Rama Chellappa.
\newblock Domain adaptation for object recognition: An unsupervised approach.
\newblock In {\em 2011 international conference on computer vision}, pages
  999--1006. IEEE, 2011.

\bibitem{graham20183d}
Benjamin Graham, Martin Engelcke, and Laurens van~der Maaten.
\newblock 3d semantic segmentation with submanifold sparse convolutional
  networks.
\newblock In {\em Proceedings of the IEEE conference on computer vision and
  pattern recognition}, pages 9224--9232, 2018.

\bibitem{graham2017submanifold}
Benjamin Graham and Laurens van~der Maaten.
\newblock Submanifold sparse convolutional networks.
\newblock {\em arXiv preprint arXiv:1706.01307}, 2017.

\bibitem{han2017high}
Xiaoguang Han, Zhen Li, Haibin Huang, Evangelos Kalogerakis, and Yizhou Yu.
\newblock High-resolution shape completion using deep neural networks for
  global structure and local geometry inference.
\newblock In {\em Proceedings of the IEEE International Conference on Computer
  Vision}, pages 85--93, 2017.

\bibitem{hanocka2019meshcnn}
Rana Hanocka, Amir Hertz, Noa Fish, Raja Giryes, Shachar Fleishman, and Daniel
  Cohen-Or.
\newblock Meshcnn: a network with an edge.
\newblock {\em ACM Transactions on Graphics (TOG)}, 38(4):1--12, 2019.

\bibitem{hoffman2017cycada}
Judy Hoffman, Eric Tzeng, Taesung Park, Jun-Yan Zhu, Phillip Isola, Kate
  Saenko, Alexei~A Efros, and Trevor Darrell.
\newblock Cycada: Cycle-consistent adversarial domain adaptation.
\newblock {\em arXiv preprint arXiv:1711.03213}, 2017.

\bibitem{huang2019texturenet}
Jingwei Huang, Haotian Zhang, Li Yi, Thomas Funkhouser, Matthias Nie{\ss}ner,
  and Leonidas~J Guibas.
\newblock Texturenet: Consistent local parametrizations for learning from
  high-resolution signals on meshes.
\newblock In {\em Proceedings of the IEEE Conference on Computer Vision and
  Pattern Recognition}, pages 4440--4449, 2019.

\bibitem{huang2018apolloscape}
Xinyu Huang, Xinjing Cheng, Qichuan Geng, Binbin Cao, Dingfu Zhou, Peng Wang,
  Yuanqing Lin, and Ruigang Yang.
\newblock The apolloscape dataset for autonomous driving.
\newblock In {\em Proceedings of the IEEE Conference on Computer Vision and
  Pattern Recognition Workshops}, pages 954--960, 2018.

\bibitem{kazhdan2006poisson}
Michael Kazhdan, Matthew Bolitho, and Hugues Hoppe.
\newblock Poisson surface reconstruction.
\newblock In {\em Proceedings of the fourth Eurographics symposium on Geometry
  processing}, volume~7, 2006.

\bibitem{lyft2019}
R. Kesten, M. Usman, J. Houston, T. Pandya, K. Nadhamuni, A. Ferreira, M. Yuan,
  B. Low, A. Jain, P. Ondruska, S. Omari, S. Shah, A. Kulkarni, A. Kazakova, C.
  Tao, L. Platinsky, W. Jiang, and V. Shet.
\newblock Lyft level 5 av dataset 2019.
\newblock url{https://level5.lyft.com/dataset/}, 2019.

\bibitem{ledig2017photo}
Christian Ledig, Lucas Theis, Ferenc Husz{\'a}r, Jose Caballero, Andrew
  Cunningham, Alejandro Acosta, Andrew Aitken, Alykhan Tejani, Johannes Totz,
  Zehan Wang, et~al.
\newblock Photo-realistic single image super-resolution using a generative
  adversarial network.
\newblock In {\em Proceedings of the IEEE conference on computer vision and
  pattern recognition}, pages 4681--4690, 2017.

\bibitem{lee2019sliced}
Chen-Yu Lee, Tanmay Batra, Mohammad~Haris Baig, and Daniel Ulbricht.
\newblock Sliced wasserstein discrepancy for unsupervised domain adaptation.
\newblock In {\em Proceedings of the IEEE Conference on Computer Vision and
  Pattern Recognition}, pages 10285--10295, 2019.

\bibitem{li2019pu}
Ruihui Li, Xianzhi Li, Chi-Wing Fu, Daniel Cohen-Or, and Pheng-Ann Heng.
\newblock Pu-gan: a point cloud upsampling adversarial network.
\newblock In {\em Proceedings of the IEEE International Conference on Computer
  Vision}, pages 7203--7212, 2019.

\bibitem{li2018pointcnn}
Yangyan Li, Rui Bu, Mingchao Sun, Wei Wu, Xinhan Di, and Baoquan Chen.
\newblock Pointcnn: Convolution on x-transformed points.
\newblock In {\em Advances in neural information processing systems}, pages
  820--830, 2018.

\bibitem{open-domain-2}
Ziwei Liu, Zhongqi Miao, Xingang Pan, Xiaohang Zhan, Stella~X Yu, Dahua Lin,
  and Boqing Gong.
\newblock Compound domain adaptation in an open world.
\newblock {\em arXiv preprint arXiv:1909.03403}, 2019.

\bibitem{long2015learning}
Mingsheng Long, Yue Cao, Jianmin Wang, and Michael~I Jordan.
\newblock Learning transferable features with deep adaptation networks.
\newblock {\em arXiv preprint arXiv:1502.02791}, 2015.

\bibitem{newell2016stacked}
Alejandro Newell, Kaiyu Yang, and Jia Deng.
\newblock Stacked hourglass networks for human pose estimation.
\newblock In {\em European conference on computer vision}, pages 483--499.
  Springer, 2016.

\bibitem{open-domain-1}
Pau Panareda~Busto and Juergen Gall.
\newblock Open set domain adaptation.
\newblock In {\em Proceedings of the IEEE International Conference on Computer
  Vision}, pages 754--763, 2017.

\bibitem{patel2015visual}
Vishal~M Patel, Raghuraman Gopalan, Ruonan Li, and Rama Chellappa.
\newblock Visual domain adaptation: A survey of recent advances.
\newblock {\em IEEE signal processing magazine}, 32(3):53--69, 2015.

\bibitem{qi2017pointnet}
Charles~R Qi, Hao Su, Kaichun Mo, and Leonidas~J Guibas.
\newblock Pointnet: Deep learning on point sets for 3d classification and
  segmentation.
\newblock In {\em Proceedings of the IEEE conference on computer vision and
  pattern recognition}, pages 652--660, 2017.

\bibitem{qi2016volumetric}
Charles~R Qi, Hao Su, Matthias Nie{\ss}ner, Angela Dai, Mengyuan Yan, and
  Leonidas~J Guibas.
\newblock Volumetric and multi-view cnns for object classification on 3d data.
\newblock In {\em Proceedings of the IEEE conference on computer vision and
  pattern recognition}, pages 5648--5656, 2016.

\bibitem{qi2017pointnet++}
Charles~Ruizhongtai Qi, Li Yi, Hao Su, and Leonidas~J Guibas.
\newblock Pointnet++: Deep hierarchical feature learning on point sets in a
  metric space.
\newblock In {\em Advances in neural information processing systems}, pages
  5099--5108, 2017.

\bibitem{qin2019pointdan}
Can Qin, Haoxuan You, Lichen Wang, C-C~Jay Kuo, and Yun Fu.
\newblock Pointdan: A multi-scale 3d domain adaption network for point cloud
  representation.
\newblock In {\em Advances in Neural Information Processing Systems}, pages
  7190--7201, 2019.

\bibitem{riegler2017octnetfusion}
Gernot Riegler, Ali~Osman Ulusoy, Horst Bischof, and Andreas Geiger.
\newblock Octnetfusion: Learning depth fusion from data.
\newblock In {\em 2017 International Conference on 3D Vision (3DV)}, pages
  57--66. IEEE, 2017.

\bibitem{rist2019cross}
Christoph~B Rist, Markus Enzweiler, and Dariu~M Gavrila.
\newblock Cross-sensor deep domain adaptation for lidar detection and
  segmentation.
\newblock In {\em 2019 IEEE Intelligent Vehicles Symposium (IV)}, pages
  1535--1542. IEEE, 2019.

\bibitem{saito2018maximum}
Kuniaki Saito, Kohei Watanabe, Yoshitaka Ushiku, and Tatsuya Harada.
\newblock Maximum classifier discrepancy for unsupervised domain adaptation.
\newblock In {\em Proceedings of the IEEE Conference on Computer Vision and
  Pattern Recognition}, pages 3723--3732, 2018.

\bibitem{saleh2019domain}
Khaled Saleh, Ahmed Abobakr, Mohammed Attia, Julie Iskander, Darius Nahavandi,
  Mohammed Hossny, and Saeid Nahvandi.
\newblock Domain adaptation for vehicle detection from bird's eye view lidar
  point cloud data.
\newblock In {\em Proceedings of the IEEE International Conference on Computer
  Vision Workshops}, pages 0--0, 2019.

\bibitem{shimodaira2000improving}
Hidetoshi Shimodaira.
\newblock Improving predictive inference under covariate shift by weighting the
  log-likelihood function.
\newblock {\em Journal of statistical planning and inference}, 90(2):227--244,
  2000.

\bibitem{shrivastava2017learning}
Ashish Shrivastava, Tomas Pfister, Oncel Tuzel, Joshua Susskind, Wenda Wang,
  and Russell Webb.
\newblock Learning from simulated and unsupervised images through adversarial
  training.
\newblock In {\em Proceedings of the IEEE conference on computer vision and
  pattern recognition}, pages 2107--2116, 2017.

\bibitem{song2017semantic}
Shuran Song, Fisher Yu, Andy Zeng, Angel~X Chang, Manolis Savva, and Thomas
  Funkhouser.
\newblock Semantic scene completion from a single depth image.
\newblock In {\em Proceedings of the IEEE Conference on Computer Vision and
  Pattern Recognition}, pages 1746--1754, 2017.

\bibitem{su2018splatnet}
Hang Su, Varun Jampani, Deqing Sun, Subhransu Maji, Evangelos Kalogerakis,
  Ming-Hsuan Yang, and Jan Kautz.
\newblock Splatnet: Sparse lattice networks for point cloud processing.
\newblock In {\em Proceedings of the IEEE Conference on Computer Vision and
  Pattern Recognition}, pages 2530--2539, 2018.

\bibitem{sugiyama2008direct}
Masashi Sugiyama, Taiji Suzuki, Shinichi Nakajima, Hisashi Kashima, Paul von
  B{\"u}nau, and Motoaki Kawanabe.
\newblock Direct importance estimation for covariate shift adaptation.
\newblock {\em Annals of the Institute of Statistical Mathematics},
  60(4):699--746, 2008.

\bibitem{sun2016deep}
Baochen Sun and Kate Saenko.
\newblock Deep coral: Correlation alignment for deep domain adaptation.
\newblock In {\em European conference on computer vision}, pages 443--450.
  Springer, 2016.

\bibitem{sun2019scalability}
Pei Sun, Henrik Kretzschmar, Xerxes Dotiwalla, Aurelien Chouard, Vijaysai
  Patnaik, Paul Tsui, James Guo, Yin Zhou, Yuning Chai, Benjamin Caine, et~al.
\newblock Scalability in perception for autonomous driving: Waymo open dataset.
\newblock {\em arXiv}, pages arXiv--1912, 2019.

\bibitem{tatarchenko2017octree}
Maxim Tatarchenko, Alexey Dosovitskiy, and Thomas Brox.
\newblock Octree generating networks: Efficient convolutional architectures for
  high-resolution 3d outputs.
\newblock In {\em Proceedings of the IEEE International Conference on Computer
  Vision}, pages 2088--2096, 2017.

\bibitem{thomas2019kpconv}
Hugues Thomas, Charles~R Qi, Jean-Emmanuel Deschaud, Beatriz Marcotegui,
  Fran{\c{c}}ois Goulette, and Leonidas~J Guibas.
\newblock Kpconv: Flexible and deformable convolution for point clouds.
\newblock In {\em Proceedings of the IEEE/CVF International Conference on
  Computer Vision}, pages 6411--6420, 2019.

\bibitem{tsai2018learning}
Yi-Hsuan Tsai, Wei-Chih Hung, Samuel Schulter, Kihyuk Sohn, Ming-Hsuan Yang,
  and Manmohan Chandraker.
\newblock Learning to adapt structured output space for semantic segmentation.
\newblock In {\em Proceedings of the IEEE Conference on Computer Vision and
  Pattern Recognition}, pages 7472--7481, 2018.

\bibitem{tzeng2017adversarial}
Eric Tzeng, Judy Hoffman, Kate Saenko, and Trevor Darrell.
\newblock Adversarial discriminative domain adaptation.
\newblock In {\em Proceedings of the IEEE Conference on Computer Vision and
  Pattern Recognition}, pages 7167--7176, 2017.

\bibitem{wang2018esrgan}
Xintao Wang, Ke Yu, Shixiang Wu, Jinjin Gu, Yihao Liu, Chao Dong, Yu Qiao, and
  Chen Change~Loy.
\newblock Esrgan: Enhanced super-resolution generative adversarial networks.
\newblock In {\em Proceedings of the European Conference on Computer Vision
  (ECCV)}, pages 0--0, 2018.

\bibitem{wang2019dynamic}
Yue Wang, Yongbin Sun, Ziwei Liu, Sanjay~E Sarma, Michael~M Bronstein, and
  Justin~M Solomon.
\newblock Dynamic graph cnn for learning on point clouds.
\newblock {\em ACM Transactions on Graphics (TOG)}, 38(5):1--12, 2019.

\bibitem{wang2019range}
Ze Wang, Sihao Ding, Ying Li, Minming Zhao, Sohini Roychowdhury, Andreas
  Wallin, Guillermo Sapiro, and Qiang Qiu.
\newblock Range adaptation for 3d object detection in lidar.
\newblock In {\em Proceedings of the IEEE International Conference on Computer
  Vision Workshops}, pages 0--0, 2019.

\bibitem{wu2019squeezesegv2}
Bichen Wu, Xuanyu Zhou, Sicheng Zhao, Xiangyu Yue, and Kurt Keutzer.
\newblock Squeezesegv2: Improved model structure and unsupervised domain
  adaptation for road-object segmentation from a lidar point cloud.
\newblock In {\em 2019 International Conference on Robotics and Automation
  (ICRA)}, pages 4376--4382. IEEE, 2019.

\bibitem{wu20153d}
Zhirong Wu, Shuran Song, Aditya Khosla, Fisher Yu, Linguang Zhang, Xiaoou Tang,
  and Jianxiong Xiao.
\newblock 3d shapenets: A deep representation for volumetric shapes.
\newblock In {\em Proceedings of the IEEE conference on computer vision and
  pattern recognition}, pages 1912--1920, 2015.

\bibitem{yang20173d}
Bo Yang, Hongkai Wen, Sen Wang, Ronald Clark, Andrew Markham, and Niki Trigoni.
\newblock 3d object reconstruction from a single depth view with adversarial
  learning.
\newblock In {\em Proceedings of the IEEE International Conference on Computer
  Vision Workshops}, pages 679--688, 2017.

\bibitem{yi2017syncspeccnn}
Li Yi, Hao Su, Xingwen Guo, and Leonidas~J Guibas.
\newblock Syncspeccnn: Synchronized spectral cnn for 3d shape segmentation.
\newblock In {\em Proceedings of the IEEE Conference on Computer Vision and
  Pattern Recognition}, pages 2282--2290, 2017.

\bibitem{yu2018ec}
Lequan Yu, Xianzhi Li, Chi-Wing Fu, Daniel Cohen-Or, and Pheng-Ann Heng.
\newblock Ec-net: an edge-aware point set consolidation network.
\newblock In {\em Proceedings of the European Conference on Computer Vision
  (ECCV)}, pages 386--402, 2018.

\bibitem{yu2018pu}
Lequan Yu, Xianzhi Li, Chi-Wing Fu, Daniel Cohen-Or, and Pheng-Ann Heng.
\newblock Pu-net: Point cloud upsampling network.
\newblock In {\em Proceedings of the IEEE Conference on Computer Vision and
  Pattern Recognition}, pages 2790--2799, 2018.

\bibitem{yue2019domain}
Xiangyu Yue, Yang Zhang, Sicheng Zhao, Alberto Sangiovanni-Vincentelli, Kurt
  Keutzer, and Boqing Gong.
\newblock Domain randomization and pyramid consistency: Simulation-to-real
  generalization without accessing target domain data.
\newblock In {\em Proceedings of the IEEE International Conference on Computer
  Vision}, pages 2100--2110, 2019.

\bibitem{zhang2018efficient}
Jiahui Zhang, Hao Zhao, Anbang Yao, Yurong Chen, Li Zhang, and Hongen Liao.
\newblock Efficient semantic scene completion network with spatial group
  convolution.
\newblock In {\em Proceedings of the European Conference on Computer Vision
  (ECCV)}, pages 733--749, 2018.

\bibitem{zhang2013domain}
Kun Zhang, Bernhard Sch{\"o}lkopf, Krikamol Muandet, and Zhikun Wang.
\newblock Domain adaptation under target and conditional shift.
\newblock In {\em International Conference on Machine Learning}, pages
  819--827, 2013.

\end{thebibliography}
}

\newpage
\appendix
\setcounter{section}{0}
\def\thesection{\Alph{section}}

This document provides a list of supplemental materials to support the main paper.
\begin{itemize}
    \item \textbf{Additional Ablation Studies} - We provide additional ablation studies in a more diverse set of domain adaptation directions in Section~\ref{sec:ablation}. Specifically, we examine the correlation between scene completion and domain adaptation performance, and we also compare our method with handcrafted sampling aligning baselines.
    \item \textbf{Loss Function for Training SVCN} - We describe the loss function for training  the sparse voxel completion network (SVCN) in detail in Section~\ref{sec:svcn}.
    \item \textbf{Label Transfer to and from the Canonical Domain} - We explain how to propagate the source-domain labels to the dense, complete point clouds in the canonical domain and how to project segmentation results from the canonical domain to the target domain in Section~\ref{sec:label_transfer}.
    \item \textbf{Implementation Details} - We provide additional implementation details of our whole pipeline in Section~\ref{sec:implementation_details}.
\end{itemize}

\section{Additional Ablation Studies}
\label{sec:ablation}

To evaluate the correlation between the quality of scene completion  and the performance of domain adaptation, we provided ablation studies using different variants of our method between Waymo and nuScenes-lidarseg in the main submission. Here we provide additional domain adaptation directions including cases between nuScenes-lidarseg and SemanticKitti and between Waymo and SemanticKitti. The settings are exactly the same as Table 3 in the main submission where we replace SVCN in our method with its variants  and  report the resulting segmentation results in Table~\ref{tab:ablation_cal}. 
It again shows that better scene completion qualities lead to better domain transfer performances, indicating the importance of high-quality surface completion in our method.

\begin{table}[h]\small
\centering
\caption{The segmentation mIoU of our approach when using different scene completion methods. N denotes nuScenes-lidarseg dataset, W denotes Waymo dataset and K denotes SemanticKITTI dataset.}
\label{tab:ablation_cal}
\newcolumntype{Y}{>{\centering\arraybackslash}X}
{
\setlength{\tabcolsep}{0.2em}
\begin{tabularx}{\columnwidth}{Y|Y|Y|Y}
\toprule
    Source$\rightarrow$ Target & Ours w/o refinement & Ours w/o adv. & Ours-full \\
    \hline
    N$\rightarrow$K & 30.1 & 32.4 & \textbf{33.7}\\
    \hline
    K$\rightarrow$N & 29.6  & 30.7 & \textbf{31.6}\\
    \hline
    W$\rightarrow$K & 58.8 & 59.5 & \textbf{60.4}\\
    \hline
    K$\rightarrow$W & 50.3  & 51.2 & \textbf{52.0}\\
\bottomrule
\end{tabularx}
}
\vspace{-10pt}
\end{table}

In addition, we also provide comparisons with handcrafted sampling aligning baselines regarding more domain adaptation directions in Table~\ref{tab:simp_baselines} to complement Table 4 in the main submission. The setting is the same as Table 4 in the main submission but we also include domain adaptation results between nuScenes-lidarseg and SemanticKitti as well as those between Waymo and SemanticKitti. Notice the adaptation between nuScenes-lidarseg and SemanticKitti includes 10 categories. Our method outperforms both B1 and B2 as well as the no adaptation baseline by large margins, demonstrating the importance of our learning based approach and the SVCN network.

\begin{table}[h]\small
\centering
\caption{Comparison with handcrafted sampling aligning baselines. N denotes nuScenes-lidarseg dataset, W denotes Waymo dataset and K denotes SemanticKITTI dataset. No DA denotes no adaptation, B1 analytically downsamples or upsamples LiDAR beams, and B2 linearly interpolates LiDAR points to densify the point cloud.}
\label{tab:simp_baselines}
\newcolumntype{Y}{>{\centering\arraybackslash}X}
{
\setlength{\tabcolsep}{0.2em}
\begin{tabularx}{\columnwidth}{Y|Y|Y|Y|Y}
\toprule
     Src$\rightarrow$Tgt & No DA& B1 & B2 & Ours \\
    \hline
    N$\rightarrow$K & 23.5 & 28.1 & 26.8 & \textbf{33.7}\\
    \hline
    K$\rightarrow$N & 27.9 & 30.3 & 29.7 & \textbf{31.6}\\
    \hline
    W$\rightarrow$K & 55.0 & - & 56.6 & \textbf{60.4}\\
    \hline
    K$\rightarrow$W & 46.3 & - & 49.5 & \textbf{52.0}\\
\bottomrule
\end{tabularx}
}
\vspace{-10pt}
\end{table}

\section{Loss Function for Training SVCN}
\label{sec:svcn}
Figure 3 in the main text shows the architectures of the structure generation network and the structure refinement network, respectively. Both networks contain 7 resolution levels. For any input-output point clouds pairs, we have the ground truth voxel existence probability (0 or 1) at each of the 7 levels. In particular, we set the groud truth voxel existence probability for a voxel to be 1 if the voxel contains one or more 3D points of the output point cloud.

To train the structure generation network, we use a binary cross entropy loss $\mathcal{L}_{\text{bce}}(c_{\text{gen}}^l, \hat{c}_{\text{gen}}^l)$ between the ground truth voxel existence probability $c_{\text{gen}}^l$ and the predicted voxel existence probability $\hat{c}_{\text{gen}}^l$, leading to a loss function $\mathcal{L}_{\text{gen}}=\underset{l}{\sum}\mathcal{L}_{\text{bce}}(c_{\text{gen}}^l, \hat{c}_{\text{gen}}^l)$, where $l$ indexes the $l$-th level of the decoder.


To train the structure refinement network, we first pre-train the structure generation network and then fix it but switch to the inference mode where we use the predicted voxel existence probability to prune voxels. A binary cross entropy loss $\mathcal{L}_{\text{refine}}=\mathcal{L}_{\text{bce}}(c_{\text{refine}}^0, \hat{c}_{\text{refine}}^0)$ at level $0$ between the ground truth voxel existence probability $c_{\text{refine}}^0$ and the predicted voxel existence probability $\hat{c}_{\text{refine}}^0$ is used to supervise the network.

\textbf{Local adversarial loss to model the prior over surfaces.}
We have a strong prior on the completed scene, namely the recovered voxels should lie on 3D surfaces. Previously, researchers have investigated a lot about how to inject high level prior knowledge to get a better loss landscape and a higher model performance. Among them adversarial learning is a successful attempt \cite{ledig2017photo, wang2018esrgan, li2019pu}. Inspired by this, we introduce local adversarial learning to further inject the 3D surface prior to our SVCN. In addition to the binary cross entropy loss we mentioned before, we add adversarial losses into $\mathcal{L}_{\text{gen}}$ and $\mathcal{L}_{\text{refine}}$, which we will detail below.

We treat SVCN as a generator which could estimate for a given incomplete LiDAR point cloud its corresponding complete counterpart and output a set of voxel existence predictions $\hat{c}_{\text{gen}}^l$ and $\hat{c}_{\text{refine}}^0$ on different resolution levels. We use $g^l$ to represent a set of voxels on resolution level $l$ from a real complete scene where each voxel is associated with an existence probability $1$. Following \cite{goodfellow2014generative}, we introduce discriminator networks $D_{\text{gen}}^l$ and $D_{\text{refine}}^0$ to differentiate $\hat{c}^l$ and $g^l$, and optimize SVCN together with the discriminators in an adversarial manner.

Instead of using a global discriminator encoding the whole scene which usually contains too much information besides the surface prior and could easily introduce complex noise for learning, we use local discriminators whose receptive field is restricted. This is achieved by using fully-convolutional architectures to retain the spatial information in the discriminator. We use the same fully-convolutional architecture for discriminators on all resolution levels. Specifically, we adopt 4 convolution layers with kernel size 3 and stride 2 followed by a linear layer in the end, where the output channel numbers are $\{32, 64, 64, 128, 1\}$. We do not use batch normalization for the discriminators. We use $D(\hat{c}^l)_i$ to represent the confidence value predicted by $D$ for the generator output $\hat{c}^l$ on each output voxel $i$. Similarly we use $D(g^l)_j$ to represent the confidence value predicted by $D$ for the real samples $g^l$ on each output voxel $j$. To train a discriminator on resolution level $l$, we use binary cross entropy to classify each output voxel into either real or fake and the loss can be written as:

\begin{equation}
    \mathcal{L}_{d}^l = -\underset{i}{\sum}\text{log}(1-D(\hat{c}^l)_i)-\underset{j}{\sum}\text{log}D(g^l)_j
\end{equation}

The adversarial loss for SVCN encourages the generator to generate voxel existence predictions fooling the discriminator and can be written as $\mathcal{L}_{\text{adv}}(\hat{c}^l)=-\underset{i}{\sum}\text{log}D(\hat{c}^l)_i$ on resolution level $l$. After adding the adversarial loss into $\mathcal{L}_{\text{gen}}$ and $\mathcal{L}_{\text{refine}}$, our final loss functional for SVCN is:

\begin{equation}
    \mathcal{L}_{\text{gen}} = \underset{l}{\sum}\mathcal{L}_{\text{bce}}(c_{
    \text{gen}}^l, \hat{c}_{\text{gen}}^l)+\lambda \mathcal{L}_{\text{adv}}(\hat{c}_{
    \text{gen}}^l)
\end{equation}

\begin{equation}
    \mathcal{L}_{\text{refine}} = \mathcal{L}_{\text{bce}}(c_{
    \text{refine}}^0, \hat{c}_{\text{refine}}^0)+\lambda \mathcal{L}_{\text{adv}}(\hat{c}_{
    \text{refine}}^0)
\end{equation}

\textbf{Confidence-aware convolution in the discriminators.}
It is worth noticing that $\hat{c}^l$ contains continuous probability values lying on densely upsampled voxels. On the other hand, $g^l$ lies on voxels from real complete scenes where each voxel is associated with an existence probability $1$. Even if SVCN predicts perfect existence scores, it is still very easy for a discriminator to tell its difference from realistic scenes using sparse convolution operations. This is to say, the gradients from discriminator will not necessarily push SVCN toward better predictions, which is against our hope. To cope with this issue, we introduce confidence-aware sparse convolution operation to replace the normal sparse convolution in all the discriminators. Recall that the sparse convolution operation proposed in \cite{graham2017submanifold} resembles normal convolution operation but restricts the computation to only active sites. To be specific, assuming $a$ represents an active voxel site, $\mathcal{N}(a)$ represents its neighboring active sites. For each $b\in\mathcal{N}(a)$, $\mathbf{f}_b$ represents the corresponding input voxel features, and $W_b$ represents the corresponding convolution kernel matrix. The output feature $\mathbf{f}^{\prime}_a$ on site $a$ after sparse convolution is $\mathbf{f}^{\prime}_a = \sum_{b\in\mathcal{N}(a)}W_b \mathbf{f}_b$. In confidence-aware sparse convolution, we have an additional confidence value $c_b$ associated with each voxel $b$ ranging from 0 to 1 and the output feature after each convolution operation becomes $\mathbf{f}^{\prime}_a = \sum_{b\in\mathcal{N}(a)}c_bW_b \mathbf{f}_b$. When applying such confidence-aware sparse convolution to $\hat{c}^l$ and $g^l$, $\hat{c}^l$ and $g^l$ will act as both input features and confidence values. It can be seen that when SVCN generates perfect voxel existence probability in either 0 or 1, the discriminator using confidence-aware sparse convolution will not be able to differentiate it from realistic scenes. Therefore confidence-aware sparse convolution is more suitable for our discriminators. To further reduce the difference between $\hat{c}^l$ and $g^l$ so that trivial solutions can be avoided and learning could start smoothly, we sharpen the predicted existence probability $\hat{c}^l$ from SVCN by replacing the sigmoid activation with a sharpened sigmoid activation $s(x)=\frac{1}{1+e^{-kx}}$ where $k\ge1$ is a sharpening factor. 

\section{Label Transfer to and from the Canonical Domain}
\label{sec:label_transfer}

In order to learn a segmentation network in the canonical domain using source domain labels while being able to infer the target domain point labels, we need two operations $\text{Prop}(\cdot)$ and $\text{Proj}(\cdot)$. $\text{Prop}(\cdot)$ propagates labels $\by_i^s$ in the source domain to the canonical domain and $\text{Proj}(\cdot)$ projects predicted labels in the canonical domain back to the target domain, resulting in predicted labels $\hat{\by}_j^t$. In this work, we simply adopt nearest neighbor based $\text{Prop}(\cdot)$ and $\text{Proj}(\cdot)$ operations. To be specific, we first voxelize input source domain point clouds $\bx_i^s$ and conduct majority-voting within each voxel to determine the voxel labels, and then for each voxel we propagate its label to its nearest neighbor voxel in the SVCN output $\psi^s(\bx_i^s)$. In the loss function, we mask out voxels without any propagated labels in $\psi^s(\bx_i^s)$ during training. At inference time, we voxelize input target domain point clouds $\bx_j^t$, fetch the voxel labels from the segmentation network predictions $\phi(\psi^t(\bx_j^t))$ through nearest neighbor search, and assign the fetched label to all the points from $\bx_j^t$ within each voxel.


\section{Implementation Details}
\label{sec:implementation_details}
The structure generation network, structure refinement network and the semantic segmentation network all contain 7 levels in their encoder-decoder architecture and adopt the same number of convolution filters on different levels. The numbers of filters from level 0 to level 6 of the encoder are $(24, 24), (24, 32), (32, 48), (48, 64), (64, 80), (80, 96), (96, 112)$ where each $(\cdot)$ corresponds to one level. The numbers of filters from level 5 to level 0 of the decoder are $(112, 96), (80, 80), (64, 64), (48, 48), (32, 32), (16, 16)$. In all our experiments, we use a voxel size of $d=20\text{cm}$. To obtain the ground truth voxel existence probability $c_{\text{gen}}^l$ on level $l$ for structure generation network training, we voxelize the ground truth complete point cloud with a voxel size of $2^ld$ and the voxel existence probability is set to be $1$ for a voxel as long as there is one point falls into it.
We use only LiDAR point positions as inputs without considering the color or intensity information. While training the segmentation network, we augment the input point clouds through randomly rotating them around z-axis and randomly flipping them with respect to the x-axis and y-axis. For both SVCN and semantic segmentation network training, we use a batch size of 2. We use Adam optimizer where the momentum is set as 0.9 and 0.99. And we use an initial learning rate of $10^{-3}$, which is decayed with a factor of 0.7 after every 200k training steps. The learning rate of the discriminator for adversarial learning is set to be $10^{-4}$ initially and also decays with a factor of 0.7 after every 200k training steps.

\end{document}